\documentclass{article} % For LaTeX2e
\usepackage[final]{colm2025_conference}

\usepackage{microtype}
\usepackage{hyperref}
\usepackage{url}
\usepackage{booktabs}

\usepackage{lineno}

\usepackage{microtype}
\usepackage{hyperref}
\usepackage{graphicx}
\usepackage{pifont}
\usepackage{multirow}
\usepackage{url}
\usepackage{booktabs}
\usepackage{amsfonts}
\usepackage{wrapfig}
\usepackage{longtable}
\newcommand{\cmark}{\ding{51}}%
\newcommand{\xmark}{\ding{55}}%

\definecolor{darkblue}{rgb}{0, 0, 0.5}
\hypersetup{colorlinks=true, citecolor=darkblue, linkcolor=darkblue, urlcolor=darkblue}

\title{debiaSAE: Benchmarking and Mitigating Vision-Language Model Bias}

\author{
 \textbf{Kuleen Sasse\textsuperscript{1,2}},
 \textbf{Shan Chen\textsuperscript{3,4,5}},
 \textbf{Jackson Pond\textsuperscript{3,4}},\\
 \textbf{Danielle S. Bitterman\textsuperscript{3,4,5*}},
 \textbf{John Osborne\textsuperscript{2}\footnotemark[1]}
\\
\\
 \textsuperscript{1}Johns Hopkins University,
 \textsuperscript{2}University of Alabama at Birmingham,
 \textsuperscript{3}Harvard Medical School,\\
 \textsuperscript{4}Mass General Brigham,
 \textsuperscript{5}Boston Children's Hospital\\\\
 \textsuperscript{*}Senior Co-Authors
\\
 \small{
   \textbf{Correspondence:} \href{mailto:ksasse@uab.edu}{ksasse@uab.edu}
 }
}

\begin{document}

\ifcolmsubmission
\linenumbers
\fi

\maketitle

\begin{abstract}

As Vision Language Models (VLMs) gain widespread use, their fairness remains under-explored. In this paper, we analyze demographic biases across five models and six datasets. We find that portrait datasets like UTKFace and CelebA are the best tools for bias detection, finding gaps in performance and fairness for both LLaVa and CLIP models. Scene-based datasets like PATA and VLStereoSet fail to be useful benchmarks for bias due to their text prompts allowing the model to guess the answer without a picture. As for pronoun-based datasets like VisoGender, we receive mixed signals as only some subsets of the data are useful in providing insights. To alleviate these two problems, we introduce a more rigorous evaluation dataset and a debiasing method based on Sparse Autoencoders to help reduce bias in models. We find that our data set generates more meaningful errors than the previous data sets. Furthermore, our debiasing method improves fairness, gaining 5-15 points in performance over the baseline. This study displays the problems with the current benchmarks for measuring demographic bias in Vision Language Models and introduces both a more effective dataset for measuring bias and a novel and interpretable debiasing method based on Sparse Autoencoders. \footnote{Code for analysis and evaluation can be found at \texttt{https://github.com/KuleenS/VLMBiasEval}} 
\end{abstract}

\section{Introduction}

Vision Language Models (VLMs) have demonstrated human and even superhuman performance across a wide range of tasks, including image captioning, visual question answering (VQA), and object detection \cite{chen-etal-2024-measuring, liu2023visual, radford2021learning}. Despite their impressive capabilities, these models exhibit harmful biases that can negatively impact user experience, perpetuate inequities, and lead to potential legal consequences \cite{kim2022race,zhang2020hurtful}. While biases and fairness in Large Language Models (LLMs) have been extensively studied, with numerous datasets created to support these efforts \cite{gallegos2023bias}, fairness and bias in VLMs have been less explored, with fewer available datasets and less thorough evaluation \cite{lee2023survey}.

In this work, we conduct a comprehensive analysis of demographic biases in VLMs, evaluating 5 models and 6 datasets used to assess those biases. We highlight the limitations of current bias evaluation datasets, demonstrating that many fail to capture the full scope of demographic bias in multimodal models. We propose future directions for developing more comprehensive and effective bias evaluation methods for VLMs, focusing on improving dataset design to mitigate conflicts between visual and textual inputs. Finally, we introduce a novel debiasing method using Sparse autoencoders (SAE) that improves fairness on two datasets across two different models. 

\begin{wrapfigure}{r}{0.5\columnwidth}
    \includegraphics[width=0.48\columnwidth]{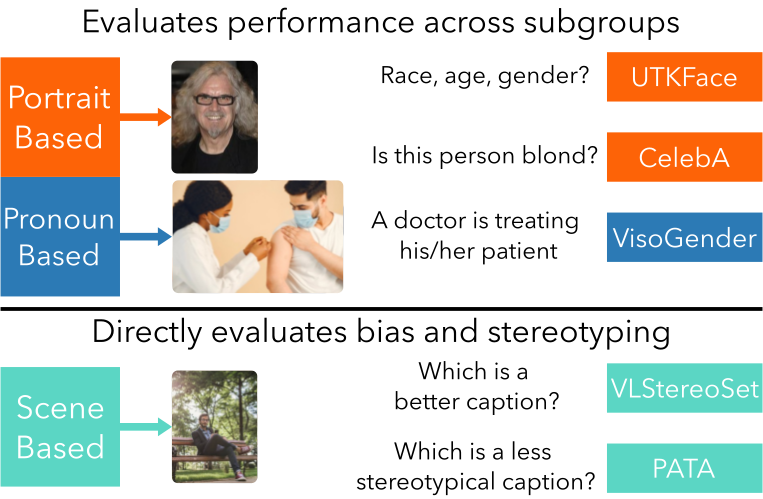}
     \caption{VLMs Demographic Bias Benchmark Taxonomy}
    \label{fig:taxonomy}
\end{wrapfigure}

\section{Related Work}
Bias in LLMs has been studied more than in VLMs, especially as these models have scaled in both capability and scope. However, as VLMs continue to evolve, their corresponding benchmarks have largely followed similar developmental frameworks as those used for LLMs. In this section, we review key studies on bias in LLMs and VLMs and categorize existing evaluation benchmarks based on biases related to portraits, scenes, and pronouns in VLMs.

\section{Related Work}
Bias in LLMs has been studied more than in VLMs, especially as these models have scaled in both capability and scope. However, as VLMs continue to evolve, their corresponding benchmarks have largely followed similar developmental frameworks as those used for LLMs. In this section, we review key studies on bias in LLMs and VLMs and categorize existing evaluation benchmarks based on biases related to portraits, scenes, and pronouns in VLMs.

\subsection{Bias in LLMs}

LLMs have made significant advancements, yet they continue to inherit and sometimes amplify societal biases, particularly in open-ended language generation tasks. Several benchmarks have been developed to quantify and mitigate these biases. For instance, the BOLD benchmark \cite{dhamala2021bold} assesses social biases across categories such as gender, race, and religion in models' free-form text generation. Similarly, \citet{parrish2022bbq} introduced the BBQ benchmark, which measures biases in multiple-choice question-answering tasks, focusing on how LLMs handle stereotypical or biased scenarios. StereoSet \cite{nadeem2020stereoset} evaluates stereotypical biases across domains such as gender, profession, race, and religion by presenting models with stereotypical and non-stereotypical sentence pairs to reveal the model’s inherent biases.

Expanding on these efforts, CrowS-Pairs \cite{nangia2020crows} evaluates bias in sentence pairs across nine categories, examining how models select between socially biased and neutral sentences. HONEST \cite{nozza2021honest} adopts a multilingual approach, analyzing harmful sentence completions to understand how LLMs generate biased or hurtful content across different languages. While these benchmarks primarily focus on text-based biases, they offer valuable insights into how LLMs may introduce similar biases when combined with visual data in VLMs.

\subsection{Bias in VLMs}

VLMs, which integrate visual and textual data, present a uniquely complex landscape for bias evaluation. Unlike purely text-based models, VLMs such as CLIP and DALL-E are susceptible to learning biases from both the images and the text used during training. This dual-source complicates and potentially amplifies fairness concerns, as prejudices from both modalities can influence the model's behavior. Several studies have explored how different pretraining methods and data mixtures impact VLM performance and bias propagation in downstream tasks \cite{wang2022unifying,fabbrizzi2023survey,cabello-etal-2023-evaluating, raj2024biasdoraexploringhiddenbiased, hamidieh2024identifyingimplicitsocialbiases}. These studies consistently highlight how VLMs can reinforce societal stereotypes related to gender, race, and professional roles.

Many datasets have been created to measure bias in VLMs, such as OccupationBias \citet{wang2022diffusiondb}, PAIRS \citet{fraser2024examining}, a synthetic dataset from \citet{sathe2024unifiedframeworkdatasetassessing}.  Another notable dataset in this area is VisoGender \citet{hall2023visogender}, which examines gender bias in pronoun resolution. VisoGender tests whether models correctly resolve pronouns in images of people performing professional roles, revealing significant gender biases in multimodal models. 

Datasets can be categorized into portrait-based, scene-based,  and the aforementioned pronoun resolution based VisoGender as shown in Figure \ref{fig:taxonomy}. Portrait based datasets like CelebA \cite{liu2015celeba} and UTKFace \cite{zhang2017utkface} are the most common to measure demographic bias in VLMs \cite{wu2024evaluatingfairnesslargevisionlanguage,han2024ffbfairfairnessbenchmark}. These datasets use portraits of peoples' faces to evaluate biases related to individual appearances, such as facial features, race, gender, and age. As for the scene-based datasets like VLStereoSet \cite{tian2022vlstereoset} and PATA \cite{singh2022pata}, these datasets evaluate how VLMs interpret complex environments involving multiple objects or people, and how bias influences their understanding of these scenes by challenging the model to choose between stereotypical and non-stereotypical captions. 

\subsection{Sparse Autoencoders: Steering and Transfer}
Sparse autoencoders have risen in popularity for use in interpreting the hidden states of large language models \cite{bricken2023towards, cunningham2023scaling}. Sparse autoencoders are shallow multilayer perceptrons with an encoder and decoder that project the hidden states of LLMs to a higher dimension sparse vector. Each dimension in this sparse vector is related to an interpretable description of what types of tokens activate that dimension. 
Because of their interpretable activations, they could be used for steering language models by intervening on those activations. Many different tasks have been shown to be effective for SAE steering, including code generation \cite{karvonen2024sieve}, chatbots \cite{goodfire}, freeform generation goals \cite{chalnev2024improvingsteeringvectorstargeting}, and refusal to answer certain unsafe prompts \cite{obrien2024steeringlanguagemodelrefusal}. 

There is very little work evaluating whether Sparse Autoencoders to can transfer to VLMs . Early work focused on transferring Sparse Autoencoders between base and chat versions of different models \cite{kutsyk2024sparse, sae_finetuning}. The only work that has successfully transferred Sparse Autoencoders to VLMs has been \cite{gallifant2025sparseautoencoderfeaturesclassifications} which has transferred Sparse Autoencoders in the context of creating classifiers for CIFAR-100 \cite{Krizhevsky09learningmultiple}. However, no prior works have been investigated to debias VLMs. 

\section{Methods}

\subsection{Datasets}
We consider five different image datasets that include protected attributes like sex, race, religion, and age to evaluate the fairness of these models: CelebA, PATA, UTKFace, Visogender, and VLStereoSet. Appendix \ref{appendix:datasets} has descriptions of the datasets with preprocessing steps. 

\subsection{Models}

We evaluate the performance and biases of various VLMs across different sizes and architectures, including both open and closed-source models. For the open-source models, we assess the LLaVa v1.6 series and the PaliGemma-2 series, specifically: \texttt{llava-v1.6-34b-hf}, \texttt{llava-v1.6-vicuna-7b}, \texttt{llava-v1.6-mistral-7b-hf} \cite{liu2023visual}, \texttt{google/paligemma2-10b-pt-224}, and \texttt{google/paligemma2-10b-pt-448} \cite{steiner2024paligemma2familyversatile}. Additionally, we evaluate two CLIP models, \texttt{CLIP L-224} and \texttt{CLIP L-336}, with the latter serving as the Vision Transformer (ViT) component for all the LLaVa models mentioned. For the closed-source models, we include Google's May 2024 release: \texttt{gemini-1.5-flash-001} \cite{radford2021learning, goldin2024gemini}. For reproducibility, we set the temperature to 0 in all cases. More details about hardware and API versioning can be found in Appendix \ref{appendix:hardware}

\begin{figure*}[ht]
    \centering
    \includegraphics[width=\textwidth]{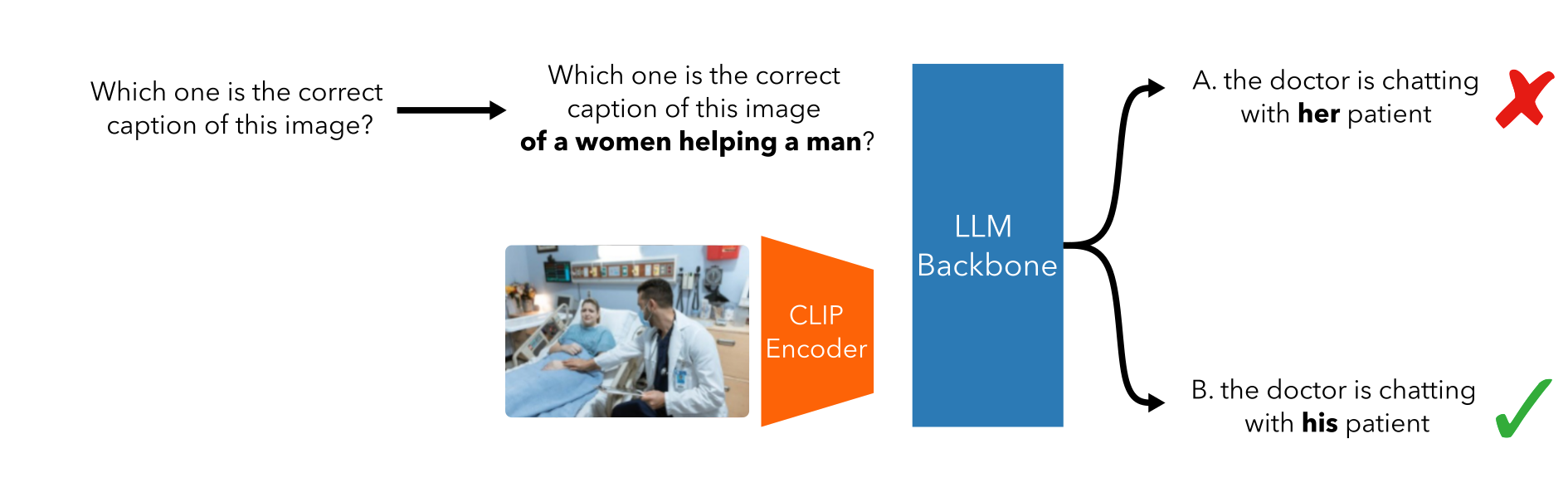}
    \caption{Diagram demonstrating our process of creating the adversarial version of the VisoGender Dataset}
    \label{fig:counter-factual}
\end{figure*}  

\subsection{Prompts and Prediction}

The prompts used in the experiments (shown in Table \ref{tab:prompts} in Appendix \ref{appendix:prompt_table}) were adapted from those recommended by the LLaVa authors for model evaluation and inspiration. The initial questions for each prompt were inspired by prompts from VQA datasets like VQAv2 \cite{goyal2017makingvvqamatter}, GQA \cite{hudson2019gqanewdatasetrealworld}, and MME \cite{fu2024mmecomprehensiveevaluationbenchmark} datasets. We manually tested and refined these prompts to ensure that the models produced accurate answers on a small subset of training splits.

For binary Yes/No questions, we ask the model the relevant question and append the recommended LLaVa prompt for binary questions: ``Answer the question using a single word or phrase.'' 

For multiple-choice questions(MCQ), we ask the model to choose the correct caption for the image from a list of options, ending with the recommended prompt for MCQ type questions for LLaVa: ``\texttt{<MCQ Options>} Answer with the option's letter from the given choices directly.''

For prediction generation, we allow models to produce a distribution over the next possible tokens. From this distribution, we select the most probable token corresponding to either ("Yes", "No") for binary questions and ("A", "B", "C", etc.) for multiple-choice questions using a process similar to the \texttt{lm-evaluation-harness} by EleutherAI \cite{eval-harness}.

To compare our models' effectiveness, we include a random classifier baseline. This baseline uniformly samples an answer from the label space for each question. We average across 100 runs to get an accurate estimate.

\subsection{Dataset Effectiveness}
Here, we define an effective dataset to measure bias as having two attributes: 1) reliance on both the visual and textual inputs and 2) difficulty. 

If the model is not given the image, the prompt alone should provide little or no information that the model could make an educated guess. This fact means for effective VLM bias benchmarking datasets, models should achieve around random or lower than random accuracy. This formulation implies that for ineffective vision-language datasets, the model will be able to guess the answer from the information provided in the prompt better than random chance. This was measured by running datasets both with and without images. 

Secondly, effective datasets should be challenging for the model. An easy dataset is ineffective at showing where the model fails. This leads to an underestimation of the bias in the model as any weakness is obscured by the overwhelming performance.  

\subsection{Adversarial Robustness}
To assess model robustness, we re-evaluate performance using an adversarial version of the VisoGender dataset, as shown in Figure \ref{fig:counter-factual}. In this experiment, we modify the prompts to intentionally mislead the model by providing incorrect gender descriptions for the individuals in the image. Specifically, we append an inaccurate gender description to the end of the question. For example, if the image shows a male doctor attending to a female patient, the modified prompt might read: "Which one is the correct caption of this image of a woman helping a man?"

Despite this adversarial input, we expect the model to identify the correct caption, such as "the doctor is chatting with his patient," rather than being misled by the incorrect gender description. This evaluation forces the model to look at the image rather than relying on any text input, as the text input is misleading.  

\subsection{Debiasing Method}
To try to remove some of the biases in the models, we propose a steering method using Sparse Autoencoders. Using the pretrained GemmaScope Sparse Autoencoders on \texttt{paligemma-2-10b-224} and \texttt{paligemma-2-10b-448}, we hand select 12 features selecting for descriptions that mentioned either 'fairness', 'gender equity', or related concepts. A table of features and their descriptions selected for the models can be found in Appendix \ref{appendix:steering_features}. 

To steer on these features, we tested five different steering methods adapted from \cite{karvonen2024sieve}: constant, conditional per input, conditional per token, clamping, and conditional clamping . \textbf{Constant} adds the Sparse Autoencoder decoder associated with the chosen feature to all tokens' hidden states scaled by a chosen coefficient. \textbf{Conditional per token} adds the vector from the SAE decoder to a token residual scaled by a chosen coefficient if that individual hidden state has an activation for the chosen feature beyond a limit. \textbf{Conditional per input} adds the vector from the SAE decoder scaled by a certain coefficient toall of the tokens' hidden states at a layer, only if at least one of the tokens' hidden states activate the chosen feature beyond a certain threshold. \textbf{Clamping} adds the decoder vector for a certain feature similar to constant, but scales the vector based on the sum of the activation of the feature and a chosen coefficient. \textbf{Conditional clamping} performing clamping when they exceed a certain limit on a feature. For each steering method, we test at different coefficients ranging from $-40$ to $40$. 

We focus on the VisoGender and Adversarial VisoGender datasets to benchmark our method. We evaluate the same as for the other datasets.

\subsection{Metrics}
We evaluate the models using three key metrics across our five datasets.

\subsubsection{Bias Dataset Effectiveness Metrics}
We measure both performance and fairness for three of the five datasets. For VisoGender and VLStereoSet, we adopt the evaluation metrics provided by those datasets.

For the portrait datasets (UTKFace, CelebA, and PATA), we use the Macro-F1 score to assess performance. To evaluate fairness, we employ the Demographic Parity Ratio (DPR) implemented in \texttt{fairlearn} \cite{bird2020fairlearn}. DPR measures the ratio between the highest and lowest selection rates. The selection rate is defined as the predicted positive rate for a given group/protected category. In a fair model, this selection rate should be independent of the underlying group/protected category, meaning the DPR should ideally be close to 100\%.

For VisoGender, we use the provided metric: Resolution Bias (RB). Resolution Bias is the difference between the number of correctly resolved pronouns for males and females within a specific occupation. A fair model should have a Resolution Bias of 0 for each occupation, indicating equal performance across genders. A positive score means that it is more accurate for females and a negative score means that it is more accurate for males. To compute the overall RB, we average the values across all occupations.

For VLStereoSet, we use the provided Vision-Language Bias Score (VLBS). VLBS is defined as the percentage of instances where the model selects a stereotypical caption over an anti-stereotypical caption. A completely unbiased model would yield a VLBS of 0.

\section{Results}

\begin{figure*}[ht]
    \centering
    \includegraphics[width=\textwidth]{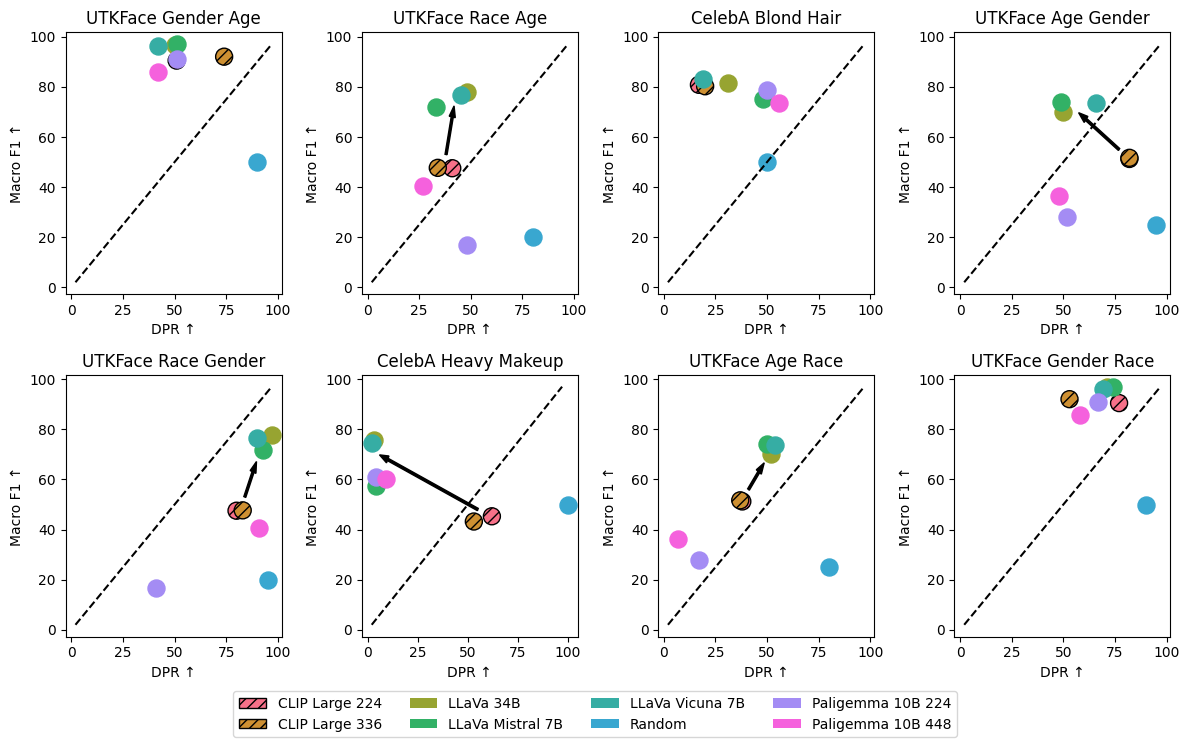}
    \caption{Performance of models on all the portrait datasets compared to their fairness (DPR demographic parity ratio, higher the better.). The dashed diagonal line represents the dividing line between trading off performance for fairness or vice versa. Arrows indicate changes in performance and fairness from CLIP to LLaVa models when the differences for both are $>$ 10\%. Note: PaliGemma did not use these CLIP based encoders. }
    \label{fig:Portraits}
\end{figure*}

\subsection{Portrait Datasets}
\paragraph{Portrait datasets exhibit key characteristics of effective datasets} Tables \ref{tab:CelebA}-\ref{tab:UTKFace-Gender} in Appendix \ref{appendix:portraittables} highlight two essential traits of effective datasets. First, including images leads to a substantial performance boost, with F1 scores increasing by 60-70 points, indicating that visual information is crucial for prediction. Second, these tasks are challenging for VLMs, as Macro F1 scores typically range from 60 to 80, with some reaching the 90s—suggesting significant room for improvement. Additionally, the Demographic Parity Ratio, which measures fairness, remains between 40 and 80, indicating that models are not yet perfectly fair under this metric, leaving room for further enhancements. 

We are able to use these datasets to make well-founded conclusions that we discuss more in Appendix \ref{appendix:analysis}. In short, adding LLMs to their base CLIP models yields more fair and more performant models. However, not all sub-tasks exhibited this improvement and therefore practitioners should be cautious when examining the overall fairness of their model using this dataset.

\subsection{Scene-Based Bias Evaluation}

\subsubsection{VLStereoSet}
\paragraph{VLStereoSet may be inadequate for studying image-induced bias in VLMs.} 
As shown in Figure \ref{fig:vlstereo} (see Appendix), VLStereoSet is ineffective at measuring bias in VLMs). Without images, models score 50-60 on the VLBS, where lower is better, and random chance yields 33. Adding images improves the scores only slightly, reducing bias by 5-15 points. This contrasts with datasets like UTKFace and CelebA, where images boost performance by 60-70 points. This means that the text provides most of the information and, therefore, models can perform well without relying on the associated imaging data.

\subsubsection{PATA}
\paragraph{PATA may have some limitations when used as a bias evaluation dataset.} 
As seen in Figure \ref{fig:PATA}, the PATA dataset also exhibits issues in its design, rendering it ineffective for evaluating bias in models. Across all models, performance either decreases or remains unchanged when images are included. This stagnation or drop in performance suggests that PATA does not require models to leverage the image input as intended. Moreover, the fairness evaluation (Figure \ref{fig:PATA}) reveals that models tend to become less fair or show no improvement when images are provided. The fact that models can achieve nearly perfect fairness using text-only inputs suggests that the inclusion of images is irrelevant and may even confuse the models, and that this dataset is ineffective for measuring bias elicited from both text and images in VLMs.

\subsection{Pronoun Resolution with VisoGender}

\paragraph{The original VisoGender dataset is insufficient to evaluate model bias.} 
In Table \ref{fig:VisoGender}, we compare the performance of our models on both the Occupation-Object and Occupation-Participant versions of the VisoGender dataset. For the Occupation-Object  subset, all models, except Gemini 1.5 Flash, achieve near-perfect accuracy. The performance is similar on the Occupation-Participant dataset, with most models maintaining around 80\% accuracy - except for Gemini 1.5 Flash, which performs worse than random. Gemini 1.5 Flash answers 95\% of the time for the female option for both datasets, likely due to interference from its safety filtering mechanisms. Despite other datasets demonstrate bias from the VLMs, as seen in UTKFace and CelebA, VisoGender but is too easy to effectively elicit errors. 

\begin{wraptable}{r}{0.6\columnwidth}
\resizebox{0.58\textwidth}{!}{
\begin{tabular}{llrrrr}
\textbf{Model}                     & \textbf{Dataset} & \multicolumn{1}{l}{\textbf{Acc. ↑}} & \multicolumn{1}{l}{\textbf{Avg. RB ↓}} & \multicolumn{1}{l}{\textbf{CF Acc. ↑}} & \multicolumn{1}{l}{\textbf{CF Avg. RB ↓}} \\ \hline
\multirow{2}{*}{Gemini 1.5 Flash}  & OO               & 49.56                               & -0.9209                                & 0.00                                   & 0.0000                                    \\
                                   & OP               & 47.33                               & -0.9397                                & 21.56                                  & -0.4266                                   \\ \hline
\multirow{2}{*}{LLaVa 34B}         & OO               & 98.68                               & -0.0179                                & 80.18                                  & -0.4148                                   \\
                                   & OP               & 87.76                               & 0.0561                                 & 57.14                                  & -0.1170                                   \\ \hline
\multirow{2}{*}{LLaVa Mistral 7B}  & OO               & 99.12                               & -0.0026                                & 92.95                                  & -0.4198                                   \\
                                   & OP               & 81.41                               & -0.2045                                & 58.50                                  & 0.0874                                    \\ \hline
\multirow{2}{*}{LLaVa Vicuna 7B}   & OO               & 96.48                               & -0.0718                                & 62.11                                  & -0.1051                                   \\
                                   & OP               & 77.10                               & -0.3431                                & 50.11                                  & -0.6981                                   \\ \hline
\multirow{2}{*}{Paligemma 10B 224} & OO               & 58.15                               & -0.7459                                & 52.42                                  & -0.8454                                   \\
                                   & OP               & 61.68                               & -0.1361                                & 50.11                                  & -0.8953                                   \\ \hline
\multirow{2}{*}{Paligemma 10B 448} & OO               & 54.63                               & -0.8907                                & 50.66                                  & -0.8299                                   \\
                                   & OP               & 57.14                               & -0.5728                                & 48.98                                  & -0.9165                                   \\ \hline
\multirow{2}{*}{Random}            & OO               & 50.00                               & 0.0000                                 & 50.00                                  & 0.0000                                    \\
                                   & OP               & 50.00                               & 0.0000                                 & 50.00                                  & 0.0000                                    \\ \hline
\end{tabular}
}
\caption{Results for VisoGender evaluation. Acc is accuracy. CF Acc is the accuracy of the adversarial version of the dataset. The performance gap between Acc vs CF Acc shows how robust models are.}

\label{fig:VisoGender}
\end{wraptable}

\subsection{Adversarial Dataset for Pronoun Resolution}

\paragraph{Adversarial prompts reveal models' reliance on textual cues over visual information.} 
As shown in Table \ref{fig:VisoGender}, model performance drops significantly when adversarial prompts are introduced to VisoGender, indicating an over-reliance on textual information even when it conflicts with visual cues. Most models experience a 20-30 point drop in accuracy from their baseline performance. Notably, LLaVa Mistral 7B shows the least degradation, with only a 6-point drop in accuracy on the Occupation-Object dataset, suggesting it is more robust to adversarial prompts than the other models. The significant accuracy drops across almost all models suggest that they tend to prioritize textual information over visual cues when the two are in conflict. This phenomenon, often referred to as "sycophancy," \cite{sharma2023understandingsycophancylanguagemodels} exposes a key weakness in current bias evaluation methods: models are highly susceptible to misleading text, which skews the bias analysis. To ensure accurate assessments of bias, datasets and prompts must be meticulously designed to avoid introducing conflicting or extraneous information.

\paragraph{The adversarial dataset serves as a more rigorous benchmark for bias evaluation.} 
The introduction of adversarial prompts shows that most models are less robust when the visual content is at odds with the textual description, underscoring the importance of well-designed datasets for bias evaluation. The adversarial version of VisoGender provides a much-needed stress test, revealing hidden biases that are not evident in the original dataset. Across most models, we see a drop in average RB on all models. In a more difficult scenario, the models shift towards defaulting towards men as their answer. This shift highlights that these models have a stronger gender bias towards males than previously predicted by the default VisoGender dataset.

\subsection{Debiasing Results}

As shown in Table \ref{tab:debias}, the best debiasing method improves performance by 3-16 points on non-adversarial VisoGender and 5-12 points on adversarial versions. This highlights the promise of SAE-based steering for debiasing and robustness against adversarial attacks. Regarding fairness, RB either decreases or remains stable in sign or magnitude. Additionally, we compared SAE debiasing to debiasing using prompts by replacing standard prompts for VisoGender from \ref{tab:prompts} with those in Appendix \ref{appendix:debiasing_prompts} as a baseline. Each prompt was handcrafted in an attempt to try to elicit a more fair response from the model.  We evaluated each prompt on the datasets. As shown in Table \ref{appendix:tab:prompting_results} (Appendix \ref{appendix:debiasing_results}), even the best debiasing prompt had no effect. This indicates that SAE-based debiasing is not only more scalable and controllable, but also a higher performing pproach than prompt-based debiasing. 

A scaling factor of 40 and a constant vector suffice for debiasing in many cases. The most impactful feature for debiasing related to "terms associated with gender dynamics and equity," aligning with the dataset's goal of assessing gender-profession separation.

\begin{table}
\resizebox{\columnwidth}{!}{
\begin{tabular}{llllllrr}
\textbf{Model}                 & \textbf{Intervention} & \textbf{Steering Method} & \textbf{Scaling Factor} & \textbf{Feature Idx.} & \textbf{Dataset}        & \multicolumn{1}{l}{\textbf{Acc. ↑}} & \multicolumn{1}{l}{\textbf{Avg. RB ↓}} \\ \hline
\multirow{8}{*}{Paligemma 224} & None                  & N/A                      & N/A                     & N/A                   & \multirow{2}{*}{OO}     & 58.15                               & -0.7459                                \\
                               & Debias                & Clamping                 & -40                     & 7624                  &                         & \textbf{69.16}                      & \textbf{-0.6006}                       \\ \cline{2-8} 
                               & None                  & N/A                      & N/A                     & N/A                   & \multirow{2}{*}{OP}     & 61.68                               & -0.1361                                \\
                               & Debias                & Clamping                 & -40                     & 7624                  &                         & \textbf{64.85}                      & \textbf{-0.0092}                       \\ \cline{2-8} 
                               & None                  & N/A                      & N/A                     & N/A                   & \multirow{2}{*}{Adv OO} & 52.42                               & -0.8454                                \\
                               & Debias                & Constant                 & 40                      & 7624                  &                         & \textbf{67.84}                      & \textbf{0.3649}                        \\ \cline{2-8} 
                               & None                  & N/A                      & N/A                     & N/A                   & \multirow{2}{*}{Adv OP} & 50.11                               & -0.8953                                \\
                               & Debias                & Constant                 & 40                      & 7624                  &                         & \textbf{55.33}                      & \textbf{0.5290}                        \\ \hline
\multirow{8}{*}{Paligemma 448} & None                  & N/A                      & N/A                     & N/A                   & \multirow{2}{*}{OO}     & 54.63                               & -0.8907                                \\
                               & Debias                & Constant                 & 40                      & 7624                  &                         & \textbf{70.93}                      & \textbf{-0.0325}                       \\ \cline{2-8} 
                               & None                  & N/A                      & N/A                     & N/A                   & \multirow{2}{*}{OP}     & 57.14                               & -0.5728                                \\
                               & Debias                & Clamping                 & -40                     & 7624                  &                         & \textbf{65.76}                      & \textbf{0.1131}                        \\ \cline{2-8} 
                               & None                  & N/A                      & N/A                     & N/A                   & \multirow{2}{*}{Adv OO} & 50.66                               & -0.8299                                \\
                               & Debias                & Constant                 & 40                      & 7624                  &                         & \textbf{62.11}                      & \textbf{-0.3463}                       \\ \cline{2-8} 
                               & None                  & N/A                      & N/A                     & N/A                   & \multirow{2}{*}{Adv OP} & 48.98                               & -0.9165                                \\
                               & Debias                & Constant                 & 40                      & 96024                 &                         & \textbf{53.51}                      & \textbf{-0.0954}                       \\ \hline
\end{tabular}
}
\caption{Results for debiasing evaluation on Visogender (OO, OP) and Adversarial Visogender (Adv OO, Adv OP). Acc is accuracy. Best debiasing method chosen for each dataset and model. Avg. RB is average resolution bias. }\label{tab:debias}
\end{table}

\section{Discussion and Conclusion}
This paper presents one of the first comprehensive evaluations of demographic bias in VLMs, analyzing state-of-the-art models across major datasets to assess model bias, the effectiveness of bias evaluation datasets, and promising debiasing methods.

\paragraph{UTKFace and CelebA datasets are effective for bias evaluation} 
UTKFace and CelebA effectively highlight fairness gaps between LLaVa and CLIP models, showing that VLMs are generally less biased and more performant when paired with an LLM backbone. However, biases in these models vary significantly depending on the dataset, emphasizing the need for diverse datasets to capture different bias dimensions.

\paragraph{VLStereoSet and PATA datasets fail to accurately measure bias} 
VLStereoSet and PATA suffer from data artifacts that allow models to exploit text cues, underestimating bias. Future work could refine these datasets to challenge models properly.

\paragraph{`Limitations of VisoGender'} 
The original VisoGender dataset is no longer an effective benchmark as models are not challenged, hiding their true biases. The adversarial version of VisoGender reveals more gender-related biases, emphasizing the potential value and need for more rigorous and carefully designed datasets that challenge models to "see" and resist being misled by textual inputs.

\paragraph{SAE Steering is Promising}
SAE based steering on the PaliGemma-2 series of models shows promise in making models less biased and more resistant to adversarial attacks. Future work can expand on implementing this steering for more datasets and models, or implementing steering at test time based on inputs. 

\section{Ethics Statement}
We identify the following limitations in our work. 
First, due to the growing popularity of multimodal models, it is impractical to evaluate all available models. We have focused on a representative subset acknowledging that this might limit the generalizability of our findings. Second, around 10 to 20 \%  of the images for VisoGender, PATA, and VLStereoSet were unable to be retrieved due to broken hyperlinks,  possibly affecting the quality of the evaluations. Third, given the limited datasets available, we recognize that the models' fairness discussed in this paper may not fully extend to real-world scenarios.

% \section*{Acknowledgments}
% % Will be ready for the camera ready version. 
% %John needs to include his grants
% The authors thank UAB Research Computing for the use of their HPC infrastructure. We acknowledge financial support from the Google PhD Fellowship (SC), the Woods Foundation (DB, SC),  the National Institute of Arthritis and Musculoskeletal and Skin Disease “Building and InnovatinG: Digital heAlth 324 Technology and Analytics (BIGDATA) 5P30AR072583”
% and the ASTRO-ACS Clinician Scientist Development Grant ASTRO-CSDG-24-1244514 (DB). The authors also thank Google Cloud research fund for Gemini API inference costs.

\bibliography{colm2024_conference}
\bibliographystyle{colm2024_conference}

\appendix
\section{Appendix}

\subsection{Dataset Description}\label{appendix:datasets}
\subsubsection{CelebA} CelebA is a large-scale portrait dataset consisting of 200k celebrities faces with 40 different attributes apparent in the portrait, such as sex, the presence of heavy makeup or glasses in the portrait. Similarly to Kim et al. \cite{10.5555/3600270.3601607}, we focus on classifying two attributes: blond hair and heavy makeup. We use the provided test set of the model to run our evaluation on. 
\subsubsection{PATA}
PATA is a set of about 5,000 public images crawled from the web that have been organized into 24 scenes. Each image is labeled with sex (male or female), race (White, Black, Indian, East-Asian, and Latino-Hispanic), and two age groups (youth and adults). Each image has two sets of captions associated with it: One set of positive captions with desirable descriptions of the person and one set of offensive or untoward text. These captions are developed with the protected attributes in mind and try to elicit a biased response from the model. For example, captions describing a person whose gender is the protected attribute could be a "photo of a smart engineer", trying to elicit a common bias that women cannot be smart and/or engineers. We use the entire dataset to run our evaluation as no training or test split is provided. 
\subsubsection{UTKFace} UTKFace is another portrait dataset that consists of over 20K portraits in the wild that cover a large variety of poses, facial expressions, illumination, etc. Each image is labeled with age, race, and sex. Similarly to \cite{yang2024demographic}, we bin the age label into 5 categories: "Child" $(0-20)$, "Young" $(20-40)$, "Middle-Aged" $(40-60)$, and "Senior" $(60-)$. We focus on classifying all three attributes and measure the bias based on the other two. We call each dataset using a two part name. The first part is what the model is predicting. The second part is what the bias is measured over. For example, "UTKFace Race Gender" would be predicting the race of the person while measuring the difference in performance across genders. As there is no test set provided, we sample 2,000 photos of the 20,000 as a test set to run our evaluation on. 
\subsubsection{VisoGender} VisoGender is a set of about 700 images that are used to benchmark gender bias in VLMs. The dataset evaluates the ability of a model to correctly resolve the pronoun of a person in a picture of them working in a professional role. The dataset measures how much the model associates different occupations with different genders. They split these images into two smaller datasets: Occupation-Object and Occupation-Participant. The Occupation-Object dataset has only one person working in a profession in the image, and the model must choose that person's pronoun. The Occupation-Participant dataset has two people in the image: one who is working in a profession and one who is participating. The task is for the model must resolve the pronoun of the one who is working. We use the entire dataset to run our prediction as no train or test split is provided. 
\subsubsection{VLStereoSet} VLStereoSet is a set of about 1000 images split between 4 categories: Gender, Profession, Race, and Religion. Each image is given both a stereotypical and anti-stereotypical caption. This dataset aims to to measure the association with a certain social group with stereotypical attributes. We use the entire dataset as no train or test split is provided.

\clearpage

\subsection{Prompt Table}\label{appendix:prompt_table}
\begin{table}[ht]
\centering
\begin{tabular}{clc}
\toprule
\textbf{Dataset} & \textbf{Prompt} & \textbf{Labels}\\
\hline
\multirow{ 2}{*}{CelebA} & Does the person in the photo have blond hair? & \multirow{ 2}{*}{Yes, No}\\
 & Does the person in the photo have heavy makeup?  & \\
\midrule
PATA & Which one is the correct caption of this image? & \multirow{6}{*}{\shortstack{Multiple\\Choice\\Question}}\\
\multirow{ 2}{*}{UTKFace} & What is the race of the person in the picture? & \\
 & What is the age of the person in the picture? & \\
 & What is the gender of the person in the picture? & \\
VisoGender & Which one is the correct caption of this image? & \\
VLStereoSet & Which one is the correct caption of this image? & \\
\bottomrule
\end{tabular}
\caption{\label{tab:prompts}
Prompt templates used in our experiments.
}
\end{table}

\clearpage

\subsection{Hardware and API Versions}\label{appendix:hardware}
All experiments were conducted on Nvidia A100 80GB GPUs, with CUDA version 12.1, transformers version 4.41.0, and the API models run through Google Vertex AI with \texttt{google-cloud-platform} version 1.60.0. 

\clearpage

\subsection{Portrait Based Dataset Tables}\label{appendix:portraittables}

% \usepackage{multirow}
% \usepackage{graphicx}

\begin{table}[htbp]
\resizebox{\columnwidth}{!}{%
\begin{tabular}{lclrr}
\textbf{Model}                     & \multicolumn{1}{l}{\textbf{Includes Image?}} & \textbf{Prediction Feature} & \multicolumn{1}{l}{\textbf{Macro F1 ↑}} & \multicolumn{1}{l}{\textbf{DPR ↑}}        \\ \hline
\multirow{4}{*}{LLaVa 34B}         & \multirow{2}{*}{\xmark}                      & Blond Hair                  & 11.76                                   & 1.00                                      \\
                                   &                                              & Heavy Makeup                & 28.82                                   & 1.00                                      \\ \cline{2-5} 
                                   & \multirow{2}{*}{\cmark}                      & Blond Hair                  & 81.47                                   & 0.31                                      \\
                                   &                                              & Heavy Makeup                & 75.77                                   & 0.03                                      \\ \hline
\multirow{4}{*}{LLaVa Mistral 7B}  & \multirow{2}{*}{\xmark}                      & Blond Hair                  & 11.76                                   & 1.00                                      \\
                                   &                                              & Heavy Makeup                & 28.82                                   & 1.00                                      \\ \cline{2-5} 
                                   & \multirow{2}{*}{\cmark}                      & Blond Hair                  & 75.19                                   & 0.48                                      \\
                                   &                                              & Heavy Makeup                & 57.51                                   & 0.04                                      \\ \hline
\multirow{4}{*}{LLaVa Vicuna 7B}   & \multirow{2}{*}{\xmark}                      & Blond Hair                  & 46.43                                   & \multicolumn{1}{r}{$\infty$} \\
                                   &                                              & Heavy Makeup                & 37.31                                   & \multicolumn{1}{r}{$\infty$} \\ \cline{2-5} 
                                   & \multirow{2}{*}{\cmark}                      & Blond Hair                  & 82.97                                   & 0.19                                      \\
                                   &                                              & Heavy Makeup                & 74.41                                   & 0.02                                      \\ \hline
\multirow{4}{*}{Paligemma 10B 224} & \multirow{2}{*}{\xmark}                      & Blond Hair                  & 46.43                                   & \multicolumn{1}{r}{$\infty$} \\
                                   &                                              & Heavy Makeup                & 37.31                                   & \multicolumn{1}{r}{$\infty$} \\ \cline{2-5} 
                                   & \multirow{2}{*}{\cmark}                      & Blond Hair                  & 78.79                                   & 0.50                                      \\
                                   &                                              & Heavy Makeup                & 60.97                                   & 0.04                                      \\ \hline
\multirow{4}{*}{Paligemma 10B 448} & \multirow{2}{*}{\xmark}                      & Blond Hair                  & 46.43                                   & \multicolumn{1}{r}{$\infty$} \\
                                   &                                              & Heavy Makeup                & 28.82                                   & 1.00                                      \\ \cline{2-5} 
                                   & \multirow{2}{*}{\cmark}                      & Blond Hair                  & 73.44                                   & 0.56                                      \\
                                   &                                              & Heavy Makeup                & 60.31                                   & 0.09                                      \\ \hline
\multirow{2}{*}{CLIP Large 224}    & \multirow{2}{*}{\cmark}                      & Blond Hair                  & 80.83                                   & 0.17                                      \\
                                   &                                              & Heavy Makeup                & 45.31                                   & 0.62                                      \\ \hline
\multirow{2}{*}{CLIP Large 336}    & \multirow{2}{*}{\cmark}                      & Blond Hair                  & 80.25                                   & 0.20                                      \\
                                   &                                              & Heavy Makeup                & 43.22                                   & 0.53                                      \\ \hline
\end{tabular}
}
\caption{Results for CelebA dataset. Predicting is the attribute that is being predicted by the model. DPR is demographic parity ratio.}\label{tab:CelebA}
\end{table}

% \usepackage{multirow}
% \usepackage{graphicx}
\begin{table}[htbp]
\resizebox{\columnwidth}{!}{%
\begin{tabular}{lclrr}
\textbf{Model}                     & \multicolumn{1}{l}{\textbf{Includes Image?}} & \textbf{Protected Category} & \multicolumn{1}{l}{\textbf{Macro F1 ↑}} & \multicolumn{1}{l}{\textbf{DPR ↑}}        \\ \hline
\multirow{4}{*}{LLaVa 34B}         & \multirow{2}{*}{\xmark}                      & Gender                      & \multirow{2}{*}{16.51}                  & 1.00                                      \\
                                   &                                              & Race                        &                                         & 1.00                                      \\ \cline{2-5} 
                                   & \multirow{2}{*}{\cmark}                      & Gender                      & \multirow{2}{*}{70.08}                  & 0.50                                      \\
                                   &                                              & Race                        &                                         & 0.52                                      \\ \hline
\multirow{4}{*}{LLaVa Mistral 7B}  & \multirow{2}{*}{\xmark}                      & Gender                      & \multirow{2}{*}{8.05}                   & \multicolumn{1}{r}{$\infty$} \\
                                   &                                              & Race                        &                                         & \multicolumn{1}{r}{$\infty$} \\ \cline{2-5} 
                                   & \multirow{2}{*}{\cmark}                      & Gender                      & \multirow{2}{*}{74.02}                  & 0.49                                      \\
                                   &                                              & Race                        &                                         & 0.50                                      \\ \hline
\multirow{4}{*}{LLaVa Vicuna 7B}   & \multirow{2}{*}{\xmark}                      & Gender                      & \multirow{2}{*}{8.76}                   & \multicolumn{1}{r}{$\infty$} \\
                                   &                                              & Race                        &                                         & \multicolumn{1}{r}{$\infty$} \\ \cline{2-5} 
                                   & \multirow{2}{*}{\cmark}                      & Gender                      & \multirow{2}{*}{73.71}                  & 0.66                                      \\
                                   &                                              & Race                        &                                         & 0.54                                      \\ \hline
\multirow{4}{*}{Paligemma 10B 224} & \multirow{2}{*}{\xmark}                      & Gender                      & \multirow{2}{*}{8.76}                   & \multicolumn{1}{r}{$\infty$} \\
                                   &                                              & Race                        &                                         & \multicolumn{1}{r}{$\infty$} \\ \cline{2-5} 
                                   & \multirow{2}{*}{\cmark}                      & Gender                      & \multirow{2}{*}{28.03}                  & 0.52                                      \\
                                   &                                              & Race                        &                                         & 0.17                                      \\ \hline
\multirow{4}{*}{Paligemma 10B 448} & \multirow{2}{*}{\xmark}                      & Gender                      & \multirow{2}{*}{8.76}                   & \multicolumn{1}{r}{$\infty$} \\
                                   &                                              & Race                        &                                         & \multicolumn{1}{r}{$\infty$} \\ \cline{2-5} 
                                   & \multirow{2}{*}{\cmark}                      & Gender                      & \multirow{2}{*}{36.29}                  & 0.48                                      \\
                                   &                                              & Race                        &                                         & 0.07                                      \\ \hline
\multirow{2}{*}{CLIP Large 224}    & \multirow{2}{*}{\cmark}                      & Gender                      & \multirow{2}{*}{51.23}                  & 0.82                                      \\
                                   &                                              & Race                        &                                         & 0.38                                      \\ \hline
\multirow{2}{*}{CLIP Large 336}    & \multirow{2}{*}{\cmark}                      & Gender                      & \multirow{2}{*}{51.68}                  & 0.82                                      \\
                                   &                                              & Race                        &                                         & 0.37                                      \\ \hline
\end{tabular}
}
\caption{Results for UTKFace Age Dataset. PC is the protected category. DPR is the demographic parity ratio.}\label{tab:UTKFace-Age}
\end{table}

% \usepackage{multirow}
% \usepackage{graphicx}
\begin{table}[htbp]
\resizebox{\columnwidth}{!}{%
\begin{tabular}{lclrr}
\textbf{Model}                     & \multicolumn{1}{l}{\textbf{Image?}} & \textbf{PC} & \multicolumn{1}{l}{\textbf{Macro F1 ↑}} & \multicolumn{1}{l}{\textbf{DPR ↑}}        \\ \hline
\multirow{4}{*}{LLaVa 34B}         & \multirow{2}{*}{\xmark}             & Age         & \multirow{2}{*}{32.45}                  & 1.00                                      \\
                                   &                                     & Race        &                                         & 1.00                                      \\ \cline{2-5} 
                                   & \multirow{2}{*}{\cmark}             & Age         & \multirow{2}{*}{96.89}                  & 0.50                                      \\
                                   &                                     & Race        &                                         & 0.71                                      \\ \hline
\multirow{4}{*}{LLaVa Mistral 7B}  & \multirow{2}{*}{\xmark}             & Age         & \multirow{2}{*}{34.20}                  & \multicolumn{1}{r}{$\infty$} \\
                                   &                                     & Race        &                                         & \multicolumn{1}{r}{$\infty$} \\ \cline{2-5} 
                                   & \multirow{2}{*}{\cmark}             & Age         & \multirow{2}{*}{97.05}                  & 0.51                                      \\
                                   &                                     & Race        &                                         & 0.74                                      \\ \hline
\multirow{4}{*}{LLaVa Vicuna 7B}   & \multirow{2}{*}{\xmark}             & Age         & \multirow{2}{*}{32.45}                  & 1.00                                      \\
                                   &                                     & Race        &                                         & 1.00                                      \\ \cline{2-5} 
                                   & \multirow{2}{*}{\cmark}             & Age         & \multirow{2}{*}{96.23}                  & 0.42                                      \\
                                   &                                     & Race        &                                         & 0.69                                      \\ \hline
\multirow{4}{*}{Paligemma 10B 224} & \multirow{2}{*}{\xmark}             & Age         & \multirow{2}{*}{34.20}                  & \multicolumn{1}{r}{$\infty$} \\
                                   &                                     & Race        &                                         & \multicolumn{1}{r}{$\infty$} \\ \cline{2-5} 
                                   & \multirow{2}{*}{\cmark}             & Age         & \multirow{2}{*}{91.06}                  & 0.51                                      \\
                                   &                                     & Race        &                                         & 0.67                                      \\ \hline
\multirow{4}{*}{Paligemma 10B 448} & \multirow{2}{*}{\xmark}             & Age         & \multirow{2}{*}{34.20}                  & \multicolumn{1}{r}{$\infty$} \\
                                   &                                     & Race        &                                         & \multicolumn{1}{r}{$\infty$} \\ \cline{2-5} 
                                   & \multirow{2}{*}{\cmark}             & Age         & \multirow{2}{*}{85.87}                  & 0.42                                      \\
                                   &                                     & Race        &                                         & 0.58                                      \\ \hline
\multirow{2}{*}{CLIP Large 224}    & \multirow{2}{*}{\cmark}             & Age         & \multirow{2}{*}{90.50}                  & 0.51                                      \\
                                   &                                     & Race        &                                         & 0.77                                      \\ \hline
\multirow{2}{*}{CLIP Large 336}    & \multirow{2}{*}{\cmark}             & Age         & \multirow{2}{*}{92.06}                  & 0.74                                      \\
                                   &                                     & Race        &                                         & 0.53                                      \\ \hline
\end{tabular}
}\caption{Results for UTKFace Gender Dataset. PC is the protected category. DPR is the demographic parity ratio.}\label{tab:UTKFace-Race}
\end{table}

% \usepackage{multirow}
% \usepackage{graphicx}
\begin{table}[htbp]
\resizebox{\columnwidth}{!}{
\begin{tabular}{lclrr}
\textbf{Model}                     & \multicolumn{1}{l}{\textbf{Includes Image?}} & \textbf{Protected Category} & \multicolumn{1}{l}{\textbf{Macro F1 ↑}} & \multicolumn{1}{l}{\textbf{DPR ↑}}        \\ \hline
\multirow{4}{*}{LLaVa 34B}         & \multirow{2}{*}{\xmark}                      & Age                         & \multirow{2}{*}{2.82}                   & \multicolumn{1}{r}{$\infty$} \\
                                   &                                              & Gender                      &                                         & \multicolumn{1}{r}{$\infty$} \\ \cline{2-5} 
                                   & \multirow{2}{*}{\cmark}                      & Age                         & \multirow{2}{*}{77.78}                  & 0.48                                      \\
                                   &                                              & Gender                      &                                         & 0.97                                      \\ \hline
\multirow{4}{*}{LLaVa Mistral 7B}  & \multirow{2}{*}{\xmark}                      & Age                         & \multirow{2}{*}{2.82}                   & \multicolumn{1}{r}{$\infty$} \\
                                   &                                              & Gender                      &                                         & \multicolumn{1}{r}{$\infty$} \\ \cline{2-5} 
                                   & \multirow{2}{*}{\cmark}                      & Age                         & \multirow{2}{*}{71.85}                  & 0.33                                      \\
                                   &                                              & Gender                      &                                         & 0.93                                      \\ \hline
\multirow{4}{*}{LLaVa Vicuna 7B}   & \multirow{2}{*}{\xmark}                      & Age                         & \multirow{2}{*}{11.97}                  & \multicolumn{1}{r}{$\infty$} \\
                                   &                                              & Gender                      &                                         & \multicolumn{1}{r}{$\infty$} \\ \cline{2-5} 
                                   & \multirow{2}{*}{\cmark}                      & Age                         & \multirow{2}{*}{76.76}                  & 0.45                                      \\
                                   &                                              & Gender                      &                                         & 0.90                                      \\ \hline
\multirow{4}{*}{Paligemma 10B 224} & \multirow{2}{*}{\xmark}                      & Age                         & \multirow{2}{*}{11.97}                  & \multicolumn{1}{r}{$\infty$} \\
                                   &                                              & Gender                      &                                         & \multicolumn{1}{r}{$\infty$} \\ \cline{2-5} 
                                   & \multirow{2}{*}{\cmark}                      & Age                         & \multirow{2}{*}{16.73}                  & 0.48                                      \\
                                   &                                              & Gender                      &                                         & 0.41                                      \\ \hline
\multirow{4}{*}{Paligemma 10B 448} & \multirow{2}{*}{\xmark}                      & Age                         & \multirow{2}{*}{11.97}                  & \multicolumn{1}{r}{$\infty$} \\
                                   &                                              & Gender                      &                                         & \multicolumn{1}{r}{$\infty$} \\ \cline{2-5} 
                                   & \multirow{2}{*}{\cmark}                      & Age                         & \multirow{2}{*}{40.58}                  & 0.27                                      \\
                                   &                                              & Gender                      &                                         & 0.91                                      \\ \hline
\multirow{2}{*}{CLIP Large 224}    & \multirow{2}{*}{\cmark}                      & Age                         & \multirow{2}{*}{47.53}                  & 0.41                                      \\
                                   &                                              & Gender                      &                                         & 0.80                                      \\ \hline
\multirow{2}{*}{CLIP Large 336}    & \multirow{2}{*}{\cmark}                      & Age                         & \multirow{2}{*}{47.68}                  & 0.34                                      \\
                                   &                                              & Gender                      &                                         & 0.83                                      \\ \hline
\end{tabular}
} \caption{Results for UTKFace Race Dataset. PC is the protected category. DPR is the demographic parity ratio.}\label{tab:UTKFace-Gender}
\end{table}

\clearpage

\subsection{Portrait Data Analysis}\label{appendix:analysis}
\paragraph{Adding the LLM on top of improves performance/fairness tradeoffs.} The results of our evaluation are presented in Figure 3. Upon closer examination, we observe a significant performance gap between the Vision Language Models (VLMs) and the baseline CLIP models. Across all datasets, the VLMs either match or outperform the CLIP models in terms of Macro-F1 score. For example, in the CelebA dataset, we see a 30-point improvement in predicting heavy makeup and a 2-point increase in predicting blondness. In the UTKFace dataset, we observe a 20-point performance boost for predicting age, a 30-point improvement for predicting race, and a modest 5-point increase for predicting gender. 

\paragraph{Fairness does not generalize consistently across tasks.} Although integrating LLMs with CLIP models generally enhances both fairness and performance for most datasets, the tradeoffs between these two factors are inconsistent. For instance, in the CelebA blond hair dataset, we observe improvements in fairness ranging from 2 to 20 points. For UTKFace Age, we see a 20-point fairness gain when measuring performance across races. For UTKFace Race, a 4-6 point improvement across ages. In UTKFace Gender, we also see a 25-point increase in fairness when measuring across races. However, these improvements are not universal. As mentioned, CelebA heavy makeup and UTKFace Age Gender exhibit a troubling reduction in fairness, raising the concern that fairness for one task does not reliably generalize to other tasks, even within the same image dataset. Practitioners should be cautious when assuming that fairness achieved in one domain will extend to others, even when they use the same image data.

\clearpage

\subsection{Scene Based Dataset Tables}
\begin{table}[ht]
\begin{tabular}{lllr}
\textbf{Model}                                             & \textbf{Image?}         & \textbf{PC} & \multicolumn{1}{l}{\textbf{VLBS ↓}} \\ \hline
                                   &                         & Gender                      & 66.67                               \\
                                   &                         & Profession                  & 58.21                               \\
                                   &                         & Race                        & 50.00                               \\
                                   & \multirow{-4}{*}{\xmark} & Religion                    & 40.00                               \\ \cline{2-4} 
                                   &                         & Gender                      & 59.69                               \\
                                   &                         & Profession                  & 54.89                               \\
                                   &                         & Race                        & 58.65                               \\
\multirow{-8}{*}{Gemini 1.5 Flash} & \multirow{-4}{*}{\cmark} & Religion                    & 20.00                               \\ \hline
                                                           &                         & Gender                      & 67.88                               \\
                                                           &                         & Profession                  & 66.92                               \\
                                                           &                         & Race                        & 61.69                               \\
                                                           & \multirow{-4}{*}{\xmark} & Religion                    & 20.00                               \\ \cline{2-4} 
                                                           &                         & Gender                      & 59.07                               \\
                                                           &                         & Profession                  & 45.38                               \\
                                                           &                         & Race                        & 53.23                               \\
\multirow{-8}{*}{LLaVa 34B}                                & \multirow{-4}{*}{\cmark} & Religion                    & 20.00                               \\ \hline
                                                           &                         & Gender                      & 74.61                               \\
                                                           &                         & Profession                  & 73.85                               \\
                                                           &                         & Race                        & 67.16                               \\
                                                           & \multirow{-4}{*}{\xmark} & Religion                    & 20.00                               \\ \cline{2-4} 
                                                           &                         & Gender                      & 59.16                               \\
                                                           &                         & Profession                  & 54.20                               \\
                                                           &                         & Race                        & 52.20                               \\
\multirow{-8}{*}{LLaVa Mistral 7B}                         & \multirow{-4}{*}{\cmark} & Religion                    & 50.00                               \\ \hline
                                                           &                         & Gender                      & 60.10                               \\
                                                           &                         & Profession                  & 63.85                               \\
                                                           &                         & Race                        & 51.74                               \\
                                                           & \multirow{-4}{*}{\xmark} & Religion                    & 20.00                               \\ \cline{2-4} 
                                                           &                         & Gender                      & 40.41                               \\
                                                           &                         & Profession                  & 41.54                               \\
                                                           &                         & Race                        & 26.87                               \\
\multirow{-8}{*}{LLaVa Vicuna 7B}                          & \multirow{-4}{*}{\cmark} & Religion                    & 20.00                               \\ \hline
\end{tabular}
\caption{Results for VLStereoSet. PC is the protected category. VLBS is the Vision-language bias score.}\label{tab:VLStereoSet}
\end{table}
\makeatletter
\setlength{\@fptop}{0pt}
\makeatother
\begin{table}[t!]
\begin{tabular}{lllr}
\textbf{Model}                                           & \textbf{Image?}         & \textbf{PC} & \multicolumn{1}{l}{\textbf{VLBS ↓}} \\ \hline
                                                         &                         & Gender                      & 97.38                               \\
                                                         &                         & Profession                  & 97.71                               \\
                                                         &                         & Race                        & 92.20                               \\
                                                         & \multirow{-4}{*}{\xmark} & Religion                    & 100.00                              \\ \cline{2-4} 
                                                         &                         & Gender                      & 78.53                               \\
                                                         &                         & Profession                  & 74.05                               \\
                                                         &                         & Race                        & 68.29                               \\
\multirow{-8}{*}{Paligemma 10B 224}                      & \multirow{-4}{*}{\cmark} & Religion                    & 66.67                               \\ \hline
                                                         &                         & Gender                      & 97.38                               \\
                                                         &                         & Profession                  & 97.71                               \\
                                                         &                         & Race                        & 92.20                               \\
                                                         & \multirow{-4}{*}{\xmark} & Religion                    & 100.00                              \\ \cline{2-4} 
                                                         &                         & Gender                      & 64.40                               \\
                                                         &                         & Profession                  & 60.31                               \\
                                                         &                         & Race                        & 60.98                               \\
\multirow{-8}{*}{Paligemma 10B 448}                      & \multirow{-4}{*}{\cmark} & Religion                    & 83.33                               \\ \hline
                                 &                         & Gender                      & 62.35                               \\
                                 &                         & Profession                  & 55.56                               \\
                                 &                         & Race                        & 64.74                               \\
\multirow{-4}{*}{CLIP Large 224} & \multirow{-4}{*}{\cmark} & Religion                    & 40.00                               \\ \hline
                                                         &                         & Gender                      & 58.64                               \\
                                                         &                         & Profession                  & 61.11                               \\
                                                         &                         & Race                        & 58.96                               \\
\multirow{-4}{*}{CLIP Large 336}                         & \multirow{-4}{*}{\cmark} & Religion                    & 40.00                               \\ \hline
\end{tabular}\caption{Results for VLStereoSet. PC is the protected category. VLBS is the Vision-language bias score.}\label{tab:VLStereoSet}
\end{table}

\begin{table}[ht]
\resizebox{\columnwidth}{!}{%
\begin{tabular}{lllrr}
\textbf{Model}                                             & \textbf{Includes Image?} & \textbf{Protected Category} & \multicolumn{1}{l}{\textbf{Macro F1 ↑}} & \multicolumn{1}{l}{\textbf{DPR ↑}} \\ \hline
                                   &                          & Age                         & 72.85                                   & 0.77                               \\
                                   &                          & Gender                      & 81.53                                   & 0.97                               \\
                                   & \multirow{-3}{*}{\xmark} & Race                        & 78.58                                   & 0.91                               \\ \cline{2-5} 
                                   &                          & Age                         & 61.72                                   & 0.79                               \\
                                   &                          & Gender                      & 85.25                                   & 0.96                               \\
\multirow{-6}{*}{Gemini 1.5 Flash} & \multirow{-3}{*}{\cmark} & Race                        & 76.26                                   & 0.90                               \\ \hline
                                                           &                          & Age                         & 92.54                                   & 0.98                               \\
                                                           &                          & Gender                      & 90.98                                   & 0.97                               \\
                                                           & \multirow{-3}{*}{\xmark} & Race                        & 94.84                                   & 0.96                               \\ \cline{2-5} 
                                                           &                          & Age                         & 95.01                                   & 0.90                               \\
                                                           &                          & Gender                      & 86.81                                   & 0.83                               \\
\multirow{-6}{*}{LLaVa 34B}                                & \multirow{-3}{*}{\cmark} & Race                        & 97.13                                   & 0.96                               \\ \hline
                                                           &                          & Age                         & 100.00                                  & 1.00                               \\
                                                           &                          & Gender                      & 98.41                                   & 1.00                               \\
                                                           & \multirow{-3}{*}{\xmark} & Race                        & 99.99                                   & 1.00                               \\ \cline{2-5} 
                                                           &                          & Age                         & 92.71                                   & 0.85                               \\
                                                           &                          & Gender                      & 92.93                                   & 0.89                               \\
\multirow{-6}{*}{LLaVa Mistral 7B}                         & \multirow{-3}{*}{\cmark} & Race                        & 95.19                                   & 0.95                               \\ \hline
                                                           &                          & Age                         & 100.00                                  & 1.00                               \\
                                                           &                          & Gender                      & 98.40                                   & 1.00                               \\
                                                           & \multirow{-3}{*}{\xmark} & Race                        & 88.65                                   & 0.98                               \\ \cline{2-5} 
                                                           &                          & Age                         & 93.27                                   & 0.88                               \\
                                                           &                          & Gender                      & 95.90                                   & 0.94                               \\
\multirow{-6}{*}{LLaVa Vicuna 7B}                          & \multirow{-3}{*}{\cmark} & Race                        & 98.35                                   & 0.96                               \\ \hline
                                                           &                          & Age                         & 98.35                                   & 0.94                               \\
                                                           &                          & Gender                      & 98.35                                   & 1.00                               \\
                                                           & \multirow{-3}{*}{\xmark} & Race                        & 98.35                                   & 0.99                               \\ \cline{2-5} 
                                                           &                          & Age                         & 94.22                                   & 0.95                               \\
                                                           &                          & Gender                      & 95.14                                   & 0.99                               \\
\multirow{-6}{*}{Paligemma 10B 224}                        & \multirow{-3}{*}{\cmark} & Race                        & 98.01                                   & 0.97                               \\ \hline
                                                           &                          & Age                         & 53.90                                   & 0.77                               \\
                                                           &                          & Gender                      & 79.50                                   & 0.97                               \\
                                                           & \multirow{-3}{*}{\xmark} & Race                        & 81.27                                   & 0.93                               \\ \cline{2-5} 
                                                           &                          & Age                         & 92.97                                   & 0.90                               \\
                                                           &                          & Gender                      & 90.02                                   & 0.99                               \\
\multirow{-6}{*}{Paligemma 10B 448}                        & \multirow{-3}{*}{\cmark} & Race                        & 90.36                                   & 0.96                               \\ \hline
                                   &                          & Age                         & 80.29                                   & 0.94                               \\
                                   &                          & Gender                      & 92.81                                   & 0.90                               \\
\multirow{-3}{*}{CLIP Large 224}   & \multirow{-3}{*}{\cmark} & Race                        & 91.26                                   & 0.96                               \\ \hline
                                                           &                          & Age                         & 80.33                                   & 0.95                               \\
                                                           &                          & Gender                      & 92.09                                   & 0.89                               \\
\multirow{-3}{*}{CLIP Large 336}                           & \multirow{-3}{*}{\cmark} & Race                        & 91.54                                   & 0.93                               \\ \hline
\end{tabular}
}\caption{Results for PATA Dataset. PC is the protected category. DPR is the demographic parity ratio.}\label{tab:PATA}
\end{table}

\clearpage

\subsection{VLStereoSet}
\begin{figure}[h]
    \centering
    \includegraphics[width=\columnwidth]{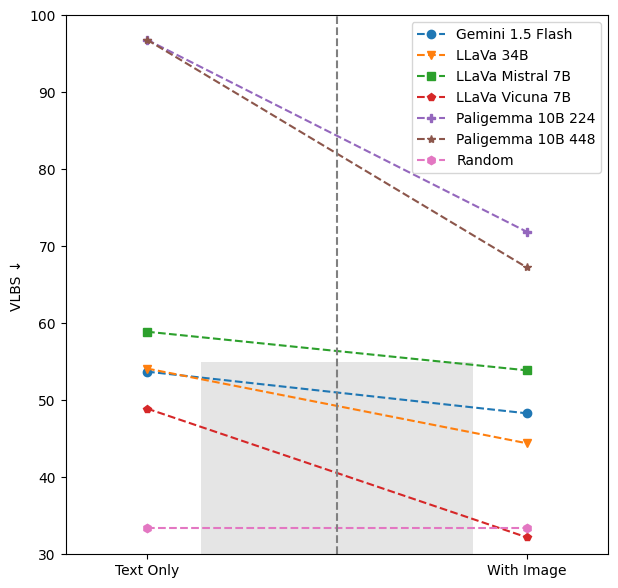}
    \caption{Comparison between "Text Only" and "With Image" of average VLBS for VLStereoSet. Grey bar represents baseline CLIP model.}
    \label{fig:vlstereo}
\end{figure}

\clearpage

\subsection{PATA}
\begin{figure*}[h]
    \centering
    \includegraphics[width=\textwidth]{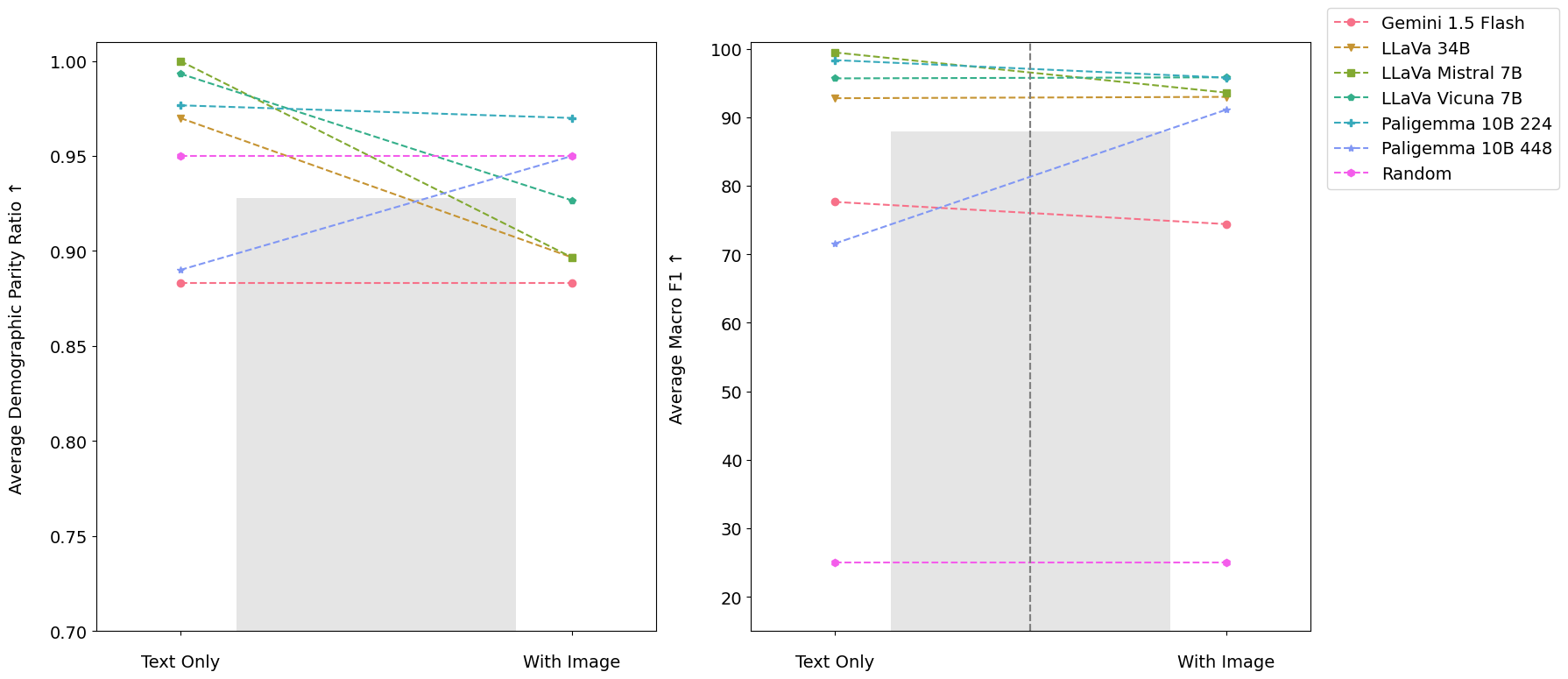}
    \caption{Average Demographic Parity Ratio comparison between "Text Only" and "With Image" for PATA (left). Macro-F1 comparison between "Text Only" and "With Image" for PATA (right). The grey bar represents the baseline CLIP Large 336 model, which is the backbone frozen vision encoder for all LLaVa that we have tested here.}
    \label{fig:PATA}
\end{figure*}

\clearpage

\subsection{Steering Features Tested}\label{appendix:steering_features}
\begin{table}[ht]
\begin{tabular}{|l|l|r|p{6cm}|}
\hline
\textbf{ID}                     & \textbf{Location} & \multicolumn{1}{l|}{\textbf{Index}} & \textbf{Description}                                                                             \\ \hline
layer\_0/width\_131k/canonical  & Residual          & 78395                               & The Concept of Fairness                                                                          \\ \hline
layer\_1/width\_131k/canonical  & MLP               & 18813                               & the concept of fairness                                                                          \\ \hline
layer\_0/width\_131k/canonical  & MLP               & 3358                                & instances of the word "fair" and its variations, indicating discussions about fairness or equity \\ \hline
layer\_35/width\_131k/canonical & Residual          & 67237                               & expressions of fairness or equity                                                                \\ \hline
layer\_16/width\_131k/canonical & MLP               & 64818                               & expressions of fairness or equity                                                                \\ \hline
layer\_18/width\_131k/canonical & MLP               & 128370                              & concepts and discussions surrounding gender equality and equality in general                     \\ \hline
layer\_30/width\_131k/canonical & MLP               & 96507                               & references to gender equality and disparities                                                    \\ \hline
layer\_35/width\_131k/canonical & Residual          & 41826                               & terms related to gender equity and pay disparity                                                 \\ \hline
layer\_13/width\_131k/canonical & MLP               & 7624                                & terms related to gender dynamics and equity                                                      \\ \hline
layer\_18/width\_131k/canonical & Residual          & 7958                                & issues related to gender equality and employment in legal and social contexts                    \\ \hline
layer\_25/width\_131k/canonical & MLP               & 96024                               & gender-associated roles and professions                                                          \\ \hline
layer\_26/width\_131k/canonical & MLP               & 19178                               & terms related to gender roles and professions                                                    \\ \hline
\end{tabular}
\caption{Information about each feature steered for Paligemma-2 10B. Features were found on Neuronpedia for Gemma-2-9B. ID is the identifier for the SAE used. It contains the layer and the width of the SAE. Location is where the hidden state was taken in the layer.}
\end{table}
\clearpage

\subsection{Prompts Used for Debiasing}\label{appendix:debiasing_prompts}
\begin{table}[h]
\begin{tabular}{cp{0.8\columnwidth}c}
\hline
\textbf{Dataset}            & \textbf{Prompt}                                                                                                                             & \textbf{Labels}                           \\ \hline
\multirow{5}{*}{VisoGender} & Based on the image which caption is most contextually appropriate while considering fairness?                                               & \multirow{5}{*}{\shortstack{Multiple\\Choice\\Question}} \\ \cline{2-2}
                            & Which caption aligns best with the image ensuring the choice is based on visual cues rather than assumptions?                               &                                           \\ \cline{2-2}
                            & Which caption best represents the image while avoiding gender-based assumptions?                                                            &                                           \\ \cline{2-2}
                            & Choose the caption that best matches the image while making a fair and unbiased decision.                                                   &                                           \\ \cline{2-2}
                            & Choose the caption that best matches the image while making a fair and unbiased decision remember that women can hold the same jobs as men. &                                           \\ \hline
\end{tabular}
\end{table}

\clearpage

\subsection{Debiasing Prompting Results}\label{appendix:debiasing_results}

\begin{table}[h]
\begin{tabular}{lllr}
\textbf{Model}                  & \textbf{Dataset}        & \textbf{Intervention} & \multicolumn{1}{l}{\textbf{Score}} \\ \hline
\multirow{12}{*}{Paligemma 224} & \multirow{3}{*}{OO}     & None                  & 0.5815                             \\
                                &                         &Best Prompt-based Debias Technique           & 0.5815                             \\
                                &                         & Best SAE-based Debias Technique           & \textbf{0.6916}                    \\ \cline{2-4} 
                                & \multirow{3}{*}{OP}     & None                  & 0.6168                             \\
                                &                         & Best Prompt-based Debias Technique           & 0.6168                             \\
                                &                         & Best SAE-based Debias Technique           & \textbf{0.6485}                    \\ \cline{2-4} 
                                & \multirow{3}{*}{Adv OO} & None                  & 0.5242                             \\
                                &                         & Best Prompt-based Debias Technique           & 0.5242                             \\
                                &                         & Best SAE-based Debias Technique           & \textbf{0.6784}                    \\ \cline{2-4} 
                                & \multirow{3}{*}{Adv OP} & None                  & 0.5011                             \\
                                &                         & Best Prompt-based Debias Technique           & 0.5011                             \\
                                &                         & Best SAE-based Debias Technique           & \textbf{0.5533}                    \\ \hline
\multirow{12}{*}{Paligemma 448} & \multirow{3}{*}{OO}     & None                  & 0.5463                             \\
                                &                         & Best Prompt-based Debias Technique           & 0.5463                             \\
                                &                         & Best SAE-based Debias Technique           & \textbf{0.7093}                    \\ \cline{2-4} 
                                & \multirow{3}{*}{OP}     & None                  & 0.5714                             \\
                                &                         & Best Prompt-based Debias Technique           & 0.5714                             \\
                                &                         & Best SAE-based Debias Technique           & \textbf{0.6576}                    \\ \cline{2-4} 
                                & \multirow{3}{*}{Adv OO} & None                  & 0.5066                             \\
                                &                         & Best Prompt-based Debias Technique           & 0.5066                             \\
                                &                         & Best SAE-based Debias Technique           & \textbf{0.6211}                    \\ \cline{2-4} 
                                & \multirow{3}{*}{Adv OP} & None                  & 0.4898                             \\
                                &                         & Best Prompt-based Debias Technique           & 0.4898                             \\
                                &                         & Best SAE-based Debias Technique           & \textbf{0.5351}                    \\ \hline
\end{tabular}\label{appendix:tab:prompting_results}
\caption{Results for debiasing evaluation on Visogender (OO, OP) and Adversarial Visogender (Adv OO, Adv OP). Acc is accuracy. Best debiasing and prompt method chosen for each dataset and model. Avg. RB is average resolution bias. }
\end{table}

\end{document}